\documentclass[10pt,twocolumn,letterpaper]{article}

\PassOptionsToPackage{table}{xcolor} 

\usepackage[pagenumbers]{wacv} 

\usepackage{booktabs}
\usepackage{multirow}
\usepackage{array}
\usepackage{tabularx}
\usepackage{amssymb}
\usepackage{pifont}
\usepackage{tikz}
\usetikzlibrary{positioning,calc,arrows.meta,patterns}
\usepackage{enumitem}
\setlist{nosep,leftmargin=*}

\newcommand{\cmark}{\ding{51}}
\newcommand{\xmark}{\textcolor{black!45}{\ding{55}}}

\newcommand{\pp}{\,pp}
\newcommand{\passk}{pass\textasciicircum$k$}
\newcommand{\para}[1]{\noindent\textbf{#1}\hspace{0.5em}}
\definecolor{lvok}{HTML}{2A78D6}
\definecolor{lvpart}{HTML}{9CC3EE}
\definecolor{lvno}{HTML}{E4E3DF}

\definecolor{wacvblue}{rgb}{0.21,0.49,0.74}
\usepackage[pagebackref,breaklinks,colorlinks,allcolors=wacvblue]{hyperref}

\def\wacvPaperID{6} 
\def\confName{WACV}
\def\confYear{2027}

\title{Trust Before Autonomy: Rethinking Agentic Remote Sensing for Crisis Response}

\author{Georgios Angelidis\\
Harokopio University of Athens\\
Athens, Greece\\
{\tt\small gangelidis@hua.gr}
\and
Aggelos Psiris\\
Harokopio University of Athens\\
Athens, Greece\\
{\tt\small aggelospsiris@hua.gr}
\and
Spyridon Evangelatos\\
Netcompany-Intrasoft S.A.\\
Luxembourg, Luxembourg\\
{\tt\small sevangelatos@netcompany.com}
\and
Thomas Lagkas\\
Dimocritus University of Thrace\\
Kavala, Greece\\
{\tt\small tlagkas@cs.duth.gr}
\and
Panagiotis Sarigiannidis\\
University of Western Macedonia\\
Kozani, Greece\\
{\tt\small psarigiannidis@uowm.gr}
\and
Iraklis Varlamis\\
Harokopio University of Athens\\
Athens, Greece\\
{\tt\small varlamis@hua.gr}
\and
Georgios Th. Papadopoulos\\
Harokopio University of Athens\\
Athens, Greece\\
{\tt\small G.Th.Papadopoulos@hua.gr}
}

\begin{document}
\maketitle

\begin{abstract}
Agentic remote sensing is evaluated on how many questions it can answer, whereas crisis response depends on whether its answers can be relied on and how much of the decision can safely be handed to the agent.
This paper argues that \emph{autonomy should be earned, not assumed}: An Earth-Observation (EO) agent should be granted autonomy in proportion to demonstrated reliability, epistemic honesty and verifiability.
The position is supported by an evidential meta-analysis of twelve agentic RS/EO/GIS benchmark studies from peer-reviewed venues.
Only one repeats runs, three contain unanswerable tasks and none measures calibration.
Their own tables show that completed workflows are often not correct (task success exceeds tool-call correctness by up to 54\pp{} on identical queries), that the gap to an expert trajectory widens overall with trajectory length (from 7 to 56\pp), and that capability does not predict abstention (Spearman $\rho$ from $-0.25$ to $0.50$, all n.s.).
The paper then contributes three EO-specific trust properties, an autonomy ladder for crisis EO tasks, and a readiness protocol that replays archived emergency activations before any live use.
In this context, current evidence justifies assistive roles only.
\end{abstract}

\section{Introduction}
\label{sec:intro}

Large language model (LLM) agents that plan, call remote-sensing tools and write code are becoming the interface to Earth Observation (EO), evolving from chains of a few vision models~\cite{guo2024rschatgpt,liu2024changeagent} to orchestrators of many tools~\cite{xu2024rsagent,earthai2025}, over a hundred in the largest~\cite{feng2026earthagent,wang2026dora}.
The first hours of a disaster require exactly the data discovery, rapid mapping and damage assessment these agents promise to automate~\cite{wang2026dora,disastercopilot2025}.

The field measures progress by \emph{capability}, the share of benchmark questions answered correctly, yet crisis response requires more.
Rapid-mapping products route rescue teams and allocate aid, and operational services subject them to quality control by the provider and an independent body~\cite{cems2024qc}.
A confident wrong map is more harmful than a late or declined one, and experts in AI for disaster management call for documentation, clearly conveyed uncertainty and independent audits~\cite{kuglitsch2024trust}.
The question is therefore not only \emph{can the agent answer?}, but \emph{when can its answer be relied on, and how much of the decision may it take alone?}

Recent position and survey papers focus on making EO agents \emph{capable}, as Munir \etal~\cite{munir2026agentic} propose trajectory-level metrics for EO as a distinct agentic domain and Talemi \etal~\cite{talemi2026agentic} provide a taxonomy of single- and multi-agent systems; neither scores uncertainty or abstention, nor asks which autonomy level the evidence supports.
Autonomy levels exist for generic agents~\cite{feng2025levels} and GIS~\cite{li2025giscience}, and the risks of full autonomy are well argued~\cite{mitchell2025fully}, but no work connects them to crisis EO or to published benchmark evidence.

\para{Position.} \emph{In crisis response, EO agents should earn autonomy step-by-step, by demonstrating reliability, epistemic honesty and verifiability; current evidence justifies only assistive roles.}
The contributions of this paper are:
\begin{itemize}
  \item An \textbf{evidential meta-analysis} of twelve agentic RS/EO/GIS benchmark studies from peer-reviewed venues (trust-coverage audit and re-analysis of their tables) that exposes a capability--trust gap (\cref{sec:evidence});
  \item Three \textbf{EO-specific trust properties} with measurable definitions (\cref{sec:trust});
  \item An \textbf{autonomy ladder} for crisis EO tasks that states the evidence each rung requires and where today's agents sit (\cref{sec:autonomy});
  \item A staged \textbf{readiness protocol} that replays archived emergency activations before any live use (\cref{sec:protocol}).
\end{itemize}

\section{Evidence: What published benchmarks reveal}
\label{sec:evidence}

\para{Method.}
The current analysis re-visits the latest versions of twelve agentic (tool-using or code-executing) RS/EO/GIS benchmark studies, published or accepted at peer-reviewed venues between 2024 and 2026, that report per-model or per-configuration results (\cref{tab:audit}).
The set comprises a selection spanning EO imagery, disaster and map-tool settings (MapEval~\cite{dihan2025mapeval} and Spatial-Agent~\cite{bao2026spatialagent} use map APIs without imagery), not a census; three are agent-method papers on shared task sets, so their evidence is not fully independent (\cref{app:limits}).
ThinkGeo~\cite{shabbir2025thinkgeo} is included as an exception, because its metrics are reused by the OpenEarthAgent~\cite{shabbir2026openearthagent} study, while preprints without acceptance are excluded.
In general, tasks and scoring differ, so absolute accuracies are \emph{never pooled}; only within-paper contrasts are analyzed (\eg step-level \vs final metrics, completion \vs correctness, run-to-run spread \vs model gaps, solvable \vs unsolvable tasks).
Rank correlations are Spearman's $\rho$ over the models of one paper and are descriptive (8--13 models).
``$\approx$'' marks values read from plotted figure positions and rounded or derived quantities.
Transcriptions, derivations and limitations are given in the appendix.

\begin{table*}[t]
\centering
\scriptsize
\setlength{\tabcolsep}{2.6pt}
\setlength{\aboverulesep}{0pt}\setlength{\belowrulesep}{0pt}
\rowcolors{2}{gray!20}{gray!2}
\begin{tabular}{l c c c c c c c c c}
\toprule
\rowcolor{gray!50} \multicolumn{1}{c}{\textbf{Benchmark}} & \textbf{Venue} & \textbf{Tasks} & \textbf{Data} & \textbf{Crisis tasks} & \textbf{Rep.\ runs} & \textbf{Unans.} & \textbf{Output scoring} & \textbf{Cost} & \textbf{Calib.} \\
\midrule
ThinkGeo~\cite{shabbir2025thinkgeo} & CVPR'26 WS$^\dagger$ & 486 & RGB, SAR & $\leq$148/436 opt.\textsuperscript{\P} & \xmark & \xmark & LLM judge & Latency & \xmark \\
Earth-Bench~\cite{feng2026earthagent} & ICLR'26 & 248 & MS, products, RGB & 24 (+13)\textsuperscript{\P} & \xmark & \xmark & Exact match (multiple choice) & Latency & \xmark \\
OpenEarthAgent~\cite{shabbir2026openearthagent} & ECCV'26 & 1{,}169 & RGB, SAR, indices, vector & $\approx$17\% & \xmark & \xmark & LLM judge & Latency & \xmark \\
DORA~\cite{wang2026dora} & NeurIPS'26$^\ast$ & 515 & 8 incl.\ SAR, DEM & all (45 events) & \xmark & \xmark & Deterministic, tolerances & Latency & \xmark \\
UnivEARTH~\cite{kao2026univearth} & ACL'26 Findings & 408 & MS, geophysical & fire $\approx$11\% & \xmark & \xmark & LLM answer extraction & -- & \xmark \\
GeoBenchX~\cite{krechetova2025geobenchx} & SIGSPATIAL'25 WS & 202 & vector, raster, tables & n.r. & \xmark & \cmark~(79) & 3-LLM judge panel & Tokens & \xmark \\
GeoNatureAgent~\cite{diazireland2026geonatureagent} & SIGSPATIAL'26$^\ast$ & 93 & rasters, statistics & -- & \cmark~(3 seeds) & \cmark & Deterministic & \$ & \xmark \\
Geo-OLM~\cite{stamoulis2025geoolm} & COMPASS'25 & 2{,}000 & RGB, MS, SAR, products & 120 (case study) & \xmark & \xmark & Deterministic$^\ddagger$ & \$, tokens & \xmark \\
GeoLLM-Engine~\cite{singh2024geollmengine} & CVPR'24 WS & 10{,}000 & RGB, MS, SAR & n.r. & \xmark & \xmark & Deterministic$^\ddagger$ & Tokens & \xmark \\
GeoLLM-Squad~\cite{lee2025geollmsquad} & IGARSS'25 & 2{,}000 & RGB, MS, SAR, products & n.r. & \xmark & \xmark & Deterministic$^\ddagger$ & Tokens & \xmark \\
Spatial-Agent~\cite{bao2026spatialagent} & ACL'26 & $\geq$547$^\S$ & map APIs, OSM & -- & \xmark & \xmark & Accuracy & \$, tokens & \xmark \\
MapEval~\cite{dihan2025mapeval} & ICML'25 & 300$^\S$ & map APIs & -- & \xmark & \cmark~(20) & Multiple choice & -- & \xmark \\
\midrule
\multicolumn{5}{l}{\textbf{Coverage of trust properties}} & \textbf{1/12} & \textbf{3/12} & \textbf{4/12} use an LLM judge & \textbf{3/12}: \$ & \textbf{0/12} \\
\bottomrule
\end{tabular}
\caption{\textbf{Trust-coverage audit} of the twelve benchmarks. $^\dagger$Non-archival workshop; cited from arXiv. $^\ast$Accepted, to appear. $^\ddagger$LLM-generated (GPT-4/4o) reference trajectories, partly human-vetted. $^\S$Map-API part only; Spatial-Agent: Minimum inferred from its tables. \textsuperscript{\P}ThinkGeo: Upper bound (includes generic change analysis); Earth-Bench: 24 disaster + 13 change detection. n.r.: Present, not counted; MS: Multispectral; products: Derived layers (\eg MODIS, GHSL); DORA latency for one model only.}
\label{tab:audit}
\vspace{-2mm}
\end{table*}

\begin{figure*}[t]
\centering
\includegraphics[width=\linewidth]{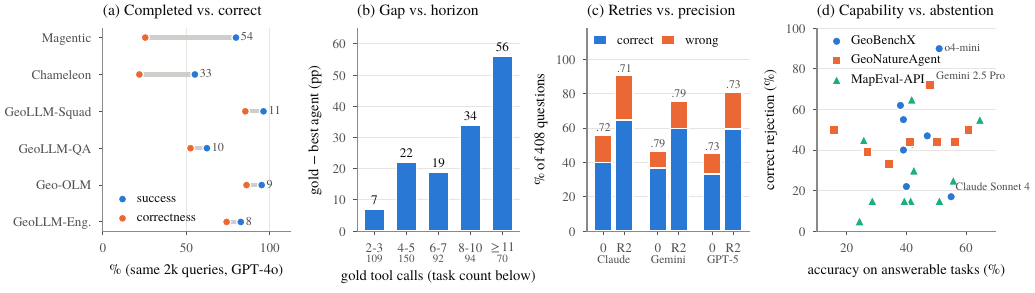}
\vspace{-6mm}
\caption{\textbf{The capability--trust gap in the benchmarks' own numbers.}
(a)~Geo-OLM, identical 2k queries, GPT-4o: Task \emph{success} \vs tool-call \emph{correctness} per framework; labels: Gap in pp~\cite{stamoulis2025geoolm}.
(b)~DORA: Accuracy gap (pp) between the gold trajectory and the best of five plotted agents, by gold length~\cite{wang2026dora}.
(c)~UnivEARTH: Correct and wrong answers at zero-shot (0) and after two reflexion rounds (R2); labels: Precision among answered questions~\cite{kao2026univearth}.
(d)~Accuracy on answerable tasks \vs correct rejection of unanswerable ones in GeoBenchX~\cite{krechetova2025geobenchx}, GeoNatureAgent~\cite{diazireland2026geonatureagent} (error handling) and MapEval-API~\cite{dihan2025mapeval}; Spearman $\rho=-0.25$ ($p=0.55$, $n=8$), $0.33$ ($p=0.38$, $n=9$) and $0.50$ ($p=0.12$, $n=11$).}
\label{fig:evidence}
\vspace{-2mm}
\end{figure*}

\para{F1. The evaluation ecosystem does not measure trust.}
Out of twelve benchmarks, one repeats runs, three contain unanswerable tasks, three report monetary cost and none elicits confidence or measures calibration (\cref{tab:audit}).
Four use an LLM in scoring (three as judges, one to extract answers), and three score against reference trajectories that an LLM generated.
The other nine, including the disaster-focused DORA, provide no credit for declining a question.

\para{F2. Right tools and completed runs are not correct answers.}
On ThinkGeo, GPT-4o's per-step tool accuracy is 67.7\%, but its end-to-end answer accuracy on optical queries is 9.8\%; across 11 models, tool accuracy does not significantly predict final answers ($\rho=0.40$, $p=0.23$; 0.69, $p=0.02$, for image-grounded ones).
On OpenEarthAgent, proprietary models reach GIS tool F1 of 86.5--95.8, but answer accuracy of only 35.2--43.9\%.
On identical queries, Geo-OLM's frameworks report task ``success'' 8.5--54.3\pp{} above tool-call correctness (Magentic: 79.7 \vs 25.4; \cref{fig:evidence}a).
Even DORA's expert gold trajectory, which uses the same perception tools, answers only 80.5\% correctly.
Earth-Bench's multiple-choice format further decouples answers from process, where LLaMA-4 scores 44.9\% (above GPT-4o's 43.7\%) with parameter accuracy of 1.3\%, and models without any tools reach about 37\%.

\para{F3. Grounding, not planning, is the bottleneck.}
Giving three DORA agents the gold tool order raises exact tool-sequence match by 25.6--28.9\pp{}, but final accuracy by only 1.1--4.4\pp.
On Earth-Bench, parameter accuracy stays at or below 31.4\% for all 13 models, while tool recall reaches 78.9\%.
DORA's failure analysis attributes 56.3\% of errors to pipeline composition (38.5\% wiring a downstream tool to the wrong upstream mask), 20.3\% to damage-semantic grounding and 14.8\% to sensor--modality mismatch.
Spatial-Agent traces 79.4\% of its analyzed failures to faulty API data or mismatched search results, none to plan construction, and 10.3\% to wrong answers after correct execution.
Retrieval itself costs accuracy, where all 11 MapEval agents score lower when they must fetch a question's context through map APIs, than when the same context is given (mean $-16.5$\pp).

\para{F4. Reliability decays with horizon and in crisis settings, and single runs cannot rank models.}
The gap between DORA's gold trajectory and the best of five plotted agents grows overall from 7\pp{} for 2--3 tool calls to 56\pp{} for 11 or more (\cref{fig:evidence}b); GeoLLM-Engine success falls from 100\% with one tool to $\approx$32--70\% with ten, depending on prompting.
DORA's top-ranked agent reaches 59.2\% on optical-only tasks, while its accuracy is lower on tasks that require vector (27.9\%), SAR (34.1\%) or multispectral+DEM+slope (17.3\%) layers, where 7 of 13 models score an identical 17.27\%, and highest among multi-modal tasks with DSM and thermal layers (53.4\%).
OpenEarthAgent's strongest models score 21.8--24.9\% on disaster assessment, among their weakest domains, \vs 39.9--53.6\% on urban and environmental tasks.
Meanwhile, the six best DORA models lie within 2.6\pp, and in GeoNatureAgent, the one benchmark with seeds, the reported spread averages $\pm$2.4\pp{} (max $\pm$4.3) and 2 of 8 adjacent rank pairs are closer than their spread.

\para{F5. Capability and honesty are different axes, and retries do not account for honesty.}
On GeoBenchX, the model best on solvable tasks (Claude Sonnet 4, 0.55) is worst at rejecting unsolvable ones (0.17), while o4-mini rejects 0.90; in GeoNatureAgent, Gemini 2.5 Pro handles 72\% of unsolvable requests \vs 50\% for the most accurate model; in MapEval-API, abstention on unanswerable questions ranges from 5\% to 65\% and is highest for an agent ranked fourth overall (\cref{fig:evidence}d).
On UnivEARTH, two rounds of reflexion raise correct answers (Claude-4.5-Sonnet 40.0$\to$64.5\%, GPT-5 33.1$\to$59.3\%), but 21--30\% of the newly produced answers are wrong and precision among answered questions stays flat (0.72$\to$0.71; 0.73$\to$0.73; \cref{fig:evidence}c).
Self-correction turns failures into answers, not into reliable answers or abstentions.

\para{F6. Scores depend on the scorer and the scaffold, and cost is rarely reported.}
GeoBenchX's judges agree with humans on 88--96\% of 50 calibration tasks; OpenEarthAgent's more lenient judge raises scores by up to $\approx$15 points without changing the ranking; in MapEval's open-ended variant, an LLM judge credits Claude with 55\% on unanswerable questions where manual inspection finds $\approx$80\%.
With the same GPT-4o-mini backbone on MapEval-API's answerable questions, accuracy ranges from $\approx$24.9\% to 45.2\% across five agent scaffolds and tool sets.
From Geo-OLM's totals, cost per unit of correctness ranges from \$0.044 to \$0.881 across frameworks with the same model; GeoNatureAgent's DeepSeek V3.2 reaches 93\% of the best model's accuracy at about 1/11 of its cost.

\section{What trust means for crisis EO}
\label{sec:trust}

Trust is defined here through three properties that an emergency service can assess, each addressing failure patterns of \cref{sec:evidence}.

\para{P1. Reliability}
(addresses F2--F4): The same query yields the same correct answer across repeated runs and across the natural variation of crisis data (sensor, region, season, haze and smoke).
It is measured with \passk~\cite{yao2024taubench}, the probability that $k$ independent runs are \emph{all} correct, estimated per task from $n\geq k$ runs with $c$ correct as $\binom{c}{k}/\binom{n}{k}$, and with accuracy on held-out events rather than held-out images.

\para{P2. Epistemic honesty}
(addresses F5): The agent abstains when the data cannot answer and reports uncertainty that survives the tool chain.
EO provides unanswerability a concrete, physical form, where the area is cloud- or smoke-covered; no post-event acquisition exists yet (revisit gap); the resolution is too coarse for the requested object; the sensor cannot observe the quantity (\eg optical imagery at night); the data are stale; or the requested class is ill-defined for the product.
Abstention is scored with selective risk--coverage~\cite{geifman2017selective,wen2025know} and abstention F1 on unanswerable variants of each task.
Uncertainty must also \emph{propagate}, where, indicatively, conformal segmentation~\cite{mossina2024conformal,mossina2025morphological,angelopoulos2023conformal} turns a flood mask into a set of possibly flooded pixels that covers the true extent with a marginal, distribution-free guarantee; carried through the tool chain, it bounds the flooded area and exposed population instead of a single number.
Calibration error~\cite{guo2017calibration} on the agent's final confidence closes the loop.

\para{P3. Verifiability}
(addresses F2 and F6): Every claim carries its evidence (scene identifiers, acquisition dates, bands, tools, parameters and intermediate masks) so that an analyst can re-run and check it.
\emph{Evidence validity} is defined as the share of claims whose cited evidence, when re-executed, reproduces the claim within the service's tolerance.
Unlike an LLM judge, it is deterministic and fits the thematic and positional checks of rapid-mapping services~\cite{cems2024qc}.

\section{Earning autonomy}
\label{sec:autonomy}

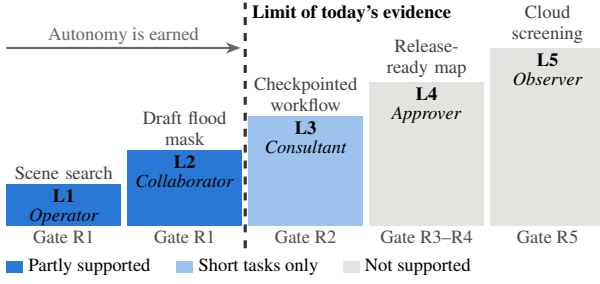
\begin{figure}[t]
\centering
\begin{tikzpicture}[x=1cm,y=1cm,font=\scriptsize]
  \foreach \i/\h/\c/\lv/\role/\task/\gate in {
      0/0.55/lvok/L1/Operator/Scene search/R1,
      1/1.00/lvok/L2/Collaborator/Draft flood mask/R1,
      2/1.45/lvpart/L3/Consultant/Checkpointed workflow/R2,
      3/1.90/lvno/L4/Approver/Release-ready map/R3--R4,
      4/2.35/lvno/L5/Observer/{Cloud screening}/R5} {
    \pgfmathsetmacro{\xl}{\i*1.6}
    \pgfmathsetmacro{\xr}{\i*1.6+1.52}
    \fill[\c] (\xl,0) rectangle (\xr,\h);
    \node[anchor=north,align=center,text width=1.45cm,inner sep=1pt] at ({\xl+0.76},{\h-0.03}) {\textbf{\lv}\\[-1pt]\textit{\role}};
    \node[anchor=south,align=center,text width=1.5cm,inner sep=1pt,text=black!75] at ({\xl+0.76},{\h+0.02}) {\task};
    \node[anchor=north,inner sep=1pt,text=black!70] at ({\xl+0.76},-0.02) {Gate \gate};
  }
  \draw[very thick,black!80,dashed] (3.16,-0.3) -- (3.16,2.95);
  \node[anchor=west,align=left,inner sep=1pt] at (3.21,2.8) {\textbf{Limit of today's evidence}};
  \draw[-{Stealth[length=2mm]},thick,black!60] (0,2.35) -- (3.1,2.35) node[midway,above,inner sep=1pt] {Autonomy is earned};
  \fill[lvok] (0,-0.62) rectangle (0.22,-0.47); \node[anchor=west,inner sep=1pt] at (0.25,-0.545) {Partly supported};
  \fill[lvpart] (2.25,-0.62) rectangle (2.47,-0.47); \node[anchor=west,inner sep=1pt] at (2.5,-0.545) {Short tasks only};
  \fill[lvno] (4.45,-0.62) rectangle (4.67,-0.47); \node[anchor=west,inner sep=1pt] at (4.7,-0.545) {Not supported};
\end{tikzpicture}
\vspace{-2mm}
\caption{\textbf{Autonomy ladder for crisis EO tasks.} Rungs follow the user roles of~\cite{feng2025levels}; above each rung an example task, below it the gate (\cref{tab:protocol}) required to reach it. Shading: Evidence of \cref{sec:evidence}.}
\label{fig:ladder}
\vspace{-2mm}
\end{figure}

Autonomy comprises a design choice, not a by-product of capability~\cite{feng2025levels,parasuraman2000model}.
\Cref{fig:ladder} adapts the five user roles of Feng \etal~\cite{feng2025levels} to crisis EO, as graded automation does for driving~\cite{sae2021j3016}, and attaches to each rung the evidence it requires.

\para{The rungs.}
At \textbf{L1 (operator)} the agent retrieves scenes and answers look-ups, and must declare when no suitable scene exists.
At \textbf{L2 (collaborator)} it drafts products (\eg a flood mask) that the analyst edits, with product accuracy and an uncertainty map.
At \textbf{L3 (consultant)} it runs a multi-step workflow with checkpoints, requiring \passk\ on replayed events and useful abstention.
At \textbf{L4 (approver)} it produces a release-ready product for human approval, requiring shadow-mode agreement with analysts and high evidence validity.
At \textbf{L5 (observer)} it acts without approval.

\para{Where agents stand.}
The capability evidence partly supports L1 and L2, where tools are often selected correctly, but retrieval and grounding errors dominate failures (F3), abstention is untested in nine of twelve benchmarks (F1, F5) and no benchmark reports uncertainty with its products.
L3 holds only for short workflows, where reliability falls beyond a few tool calls and is not separable from run-to-run noise (F4).
L4 is not supported, where completion is not correctness (F2), scores shift with the scorer (F6), and no replay or shadow-mode study of an EO agent has been published.
L5 suits only low-consequence sub-tasks such as cloud screening, since risk to people grows with autonomy~\cite{mitchell2025fully}.
No agent has formally passed even gate R1 (no benchmark reports both \passk\ and abstention); benchmarks must change before any rung above L2 is justified, so current evidence supports assistive roles (L1--L2) only.

\section{A readiness protocol}
\label{sec:protocol}

\begin{table}[t]
\centering
\scriptsize
\setlength{\tabcolsep}{2.6pt}
\renewcommand{\arraystretch}{1.08}
\setlength{\aboverulesep}{0pt}\setlength{\belowrulesep}{0pt}
\rowcolors{2}{gray!20}{gray!2}
\renewcommand{\tabularxcolumn}[1]{m{#1}}
\begin{tabularx}{\linewidth}{l >{\centering\arraybackslash\hsize=0.8\hsize}X >{\centering\arraybackslash\hsize=1.2\hsize}X >{\centering\arraybackslash}m{1.45cm}}
\toprule
\rowcolor{gray!50} \multicolumn{1}{c}{\textbf{Stage}} & \textbf{Setting} & \textbf{Measured} & \textbf{Gate to next} \\
\midrule
\textbf{R1} Benchmark & Curated tasks with unanswerable variants & Accuracy, \passk, abstention F1, risk--coverage, calibration error, \$ per correct answer & Stable \passk\ and abstention above user threshold \\
\textbf{R2} Replay & Archived activations, reference = delivered products & Agreement with reference, evidence validity, time to product & Within the service's QC tolerance \\
\textbf{R3} Shadow & Live events, nothing released & Agreement with analysts, useful-abstention rate, analyst time saved & Sustained across event types \\
\textbf{R4} Supervised & Outputs enter the normal QC chain & Rejection rate at QC, correction effort & Low rejection over a set period \\
\textbf{R5} Conditional & Low-consequence sub-tasks only & Continuous monitoring, drift & Revoked on any drift \\
\bottomrule
\end{tabularx}
\caption{\textbf{Staged readiness protocol.} Thresholds are set by the operational user; the protocol fixes the structure, not the numbers.}
\label{tab:protocol}
\vspace{-2mm}
\end{table}

Benchmarks are necessary but not sufficient, since an agent is promoted one rung of \cref{fig:ladder} only after passing the corresponding gate of \cref{tab:protocol}, and demoted on failure.
\textbf{R1} upgrades existing benchmarks cheaply, by repeating each task $k\!=\!5$ times, adding unanswerable variants built from the EO causes of P2 (a cloudy post-event scene, a sub-meter object in 10\,m data), eliciting a confidence with every answer, and reporting cost alongside accuracy, \eg dollars per correct answer~\cite{kapoor2024agents,kapoor2025hal}.
\textbf{R2}, \emph{retrospective replay}, is the most EO-specific step, where public archives of past emergency activations hold dated requests, the imagery then available and the products analysts delivered, so replay tests an agent against an operational reference at no risk to anyone.
\textbf{R3}--\textbf{R5} then move from shadow runs during live events to the existing quality-control chain~\cite{cems2024qc} and to conditional autonomy for sub-tasks, whose errors are cheap and quickly detected.

\section{Alternative views and call to action}
\label{sec:views}

\para{``Accuracy will keep rising; trust will follow.''}
The GeoBenchX model most accurate on solvable tasks rejects the fewest unsolvable ones, and retries raise coverage at constant precision (F5), where trust must be targeted directly.

\para{``A human is in the loop anyway.''}
A human can only check what is shown; without uncertainty and evidence (P2, P3), oversight of a long tool chain becomes rubber-stamping.

\para{``Abstention makes agents less useful.''}
Risk--coverage curves let each service choose its operating point; a declined request costs minutes, a confident wrong map can cost lives.

\para{``Operational services already have quality control.''}
Agents should plug into it (R4), which needs checkable evidence (P3).

\para{Call to action.}\label{sec:cta}
\textbf{Benchmark builders} should adopt R1, including \passk\ over repeated runs, unanswerable EO tasks, deterministic scoring where possible and cost per correct answer. \textbf{Agent developers} should emit confidence and evidence with every answer and propagate tool uncertainty. \textbf{Emergency services} should open archived activations for replay and grant autonomy by rung, not by leaderboard.

\section*{Use of large language models}
Claude (Anthropic) was used to extract values from the benchmark papers, compute statistics, check grammar and generate the figures; all values are traceable to the cited sources (appendix), and the authors take full responsibility for all content.

{
    \small
    \bibliographystyle{ieeenat_fullname}
    \bibliography{bibliography}
}

\clearpage
\appendix

\section*{Appendix: Meta-analysis details}
\addcontentsline{toc}{section}{Appendix}
\renewcommand{\thesubsection}{\Alph{subsection}}
\setcounter{table}{0}
\renewcommand{\thetable}{A\arabic{table}}

This appendix documents the evidential meta-analysis of \cref{sec:evidence}, namely the inclusion protocol, the values transcribed from each benchmark, the derived statistics, the ten insights behind findings F1--F6, and the limitations of the analysis.
All values are percentages unless stated otherwise; ``T.'' refers to a table and ``Fig.'' to a figure of the \emph{cited} paper.

\subsection{Protocol}
\label{app:protocol}

\para{Inclusion.} Agentic (tool-using or code-executing) benchmarks for remote sensing, EO or GIS, January 2024 to September 2026, with per-model or per-configuration result tables, published or accepted at a peer-reviewed venue (\eg ICLR, ICML, NeurIPS, ECCV, ACL, ACM SIGSPATIAL, ACM COMPASS, IEEE IGARSS, CVPR workshops).
Two map-tool benchmarks (MapEval, Spatial-Agent) are included because they evaluate geospatial tool use through map APIs; MapEval's agentic API track is analyzed, and its Textual track serves only as the context-given reference for the same questions and for the human baseline.
\para{Selection.} The twelve studies are a selection spanning EO imagery, disaster and map-tool settings; other peer-reviewed agentic GIS studies exist, so the set is not a census.
ThinkGeo is included as the only exception (non-archival CVPR 2026 workshop) because its metric suite is reused by OpenEarthAgent.
\para{Exclusion.} Preprints without acceptance (\eg GISAgentBench, GeoAgentBench, OpenEarth-Bench, GISclaw, SimCRAFT, Earth-Bench-Pro) and pure VLM benchmarks without tools.
\para{Extraction.} Figure values are used when printed on the figure; values read from plotted positions, and rounded or derived quantities, are marked $\approx$.
\para{Analysis.} No cross-benchmark pooling of absolute accuracy. Within-paper contrasts only. Spearman $\rho$ with two-sided $p$ (\texttt{scipy.stats.spearmanr}); with 8--13 models these are descriptive, not confirmatory.

\subsection{Transcribed results and derived statistics}
\label{app:data}

\para{ThinkGeo~\cite{shabbir2025thinkgeo}} (486 queries: 436 optical, 50 SAR; 14 tools; 11 models; answers scored by a GPT-4o-mini judge).
\cref{tab:a-thinkgeo} reproduces T.3 (optical).
Spearman $\rho$ against final answer accuracy (Ans), $n=11$: InstAcc 0.45 ($p=0.16$), ToolAcc 0.40 (0.23), ArgAcc 0.33 (0.32), SummAcc 0.40 (0.23); none significant.
Against image-grounded accuracy (Ans$_I$): InstAcc 0.75 ($p=0.008$), ToolAcc 0.69 (0.019), ArgAcc 0.69 (0.018).
The paper's Fig.~5 reports Pearson $r$ (ToolAcc 0.435); Pearson values recomputed from T.3 match it.
SAR (T.A1, 8 models): All Ans values are 2.78 or 5.56; GPT-4o 5.56 despite ToolAcc 77.87.
Latency (T.A4, 6 models): 6.8--27.0\,s per query.

\begin{table}[h]
\centering\scriptsize\setlength{\tabcolsep}{3pt}
\setlength{\aboverulesep}{0pt}\setlength{\belowrulesep}{0pt}
\rowcolors{2}{gray!20}{gray!2}
\begin{tabular}{lccccc}
\toprule
\rowcolor{gray!50} \multicolumn{1}{c}{\textbf{Model}} & \textbf{Inst} & \textbf{Tool} & \textbf{Arg} & \textbf{Ans} & \textbf{Ans$_I$} \\
\midrule
GPT-4o & 82.33 & 67.73 & 34.75 & 9.78 & 20.40 \\
GPT-4-1106 & 86.49 & 74.05 & 36.96 & 5.16 & 14.69 \\
Claude-Sonnet & 22.31 & 27.31 & 0.00 & 8.97 & 7.57 \\
Qwen1.5-7B & 26.92 & 11.47 & 3.06 & 5.43 & 4.59 \\
Qwen2.5-7B & 64.88 & 51.04 & 20.08 & 7.34 & 6.40 \\
InternLM3-8B & 50.53 & 45.37 & 21.38 & 8.15 & 9.15 \\
LLaMA3.1-8B & 47.27 & 37.29 & 13.69 & 3.80 & 3.42 \\
Phi-3-mini & 38.28 & 31.49 & 13.30 & 6.25 & 5.28 \\
Mistral-7B & 21.35 & 20.40 & 0.00 & 3.80 & 3.21 \\
Yi-1.5-6B & 22.29 & 23.27 & 0.20 & 3.26 & 2.75 \\
Qwen3-8B & 26.03 & 14.73 & 3.19 & 5.16 & 5.63 \\
\bottomrule
\end{tabular}
\caption{ThinkGeo, optical split (T.3 of~\cite{shabbir2025thinkgeo}).}
\label{tab:a-thinkgeo}
\end{table}

\para{OpenEarthAgent~\cite{shabbir2026openearthagent}} (1{,}169 test tasks; 24 tools; 10 models; answers scored by a GPT-4o-mini judge with $\pm10\%$ numeric tolerance).
\cref{tab:a-oea} combines T.3 (step-by-step mode: Tool, ArgV) and T.4 (end-to-end mode: GIS F1, Ans); the two modes differ, so tool--answer pairs are indicative.
Over the nine baselines, $\rho$(Tool, Ans) $=0.70$ ($p=0.036$) and $\rho$(ArgV, Ans) $=0.85$ ($p=0.004$); ArgV ranges 12.5--45.8 for baselines.
Domain heatmap (Fig.~A3): Disaster assessment 24.7 / 24.9 / 21.8 for OEA / GPT-4o / GPT-5, against urban planning 51.9 / 39.9 / 52.4 and environmental monitoring 51.5 / 53.6 / 53.2; transportation is lower still for these models (17.3--21.0), so disaster is among, not uniquely, the weakest domains.
Judge sensitivity (Fig.~A7, bars not labeled, $\approx$): Qwen3-30B \vs GPT-4o-mini adds $\approx$15 points for the fine-tuned model, $\approx$7--9.5 for GPT-4o, o4-mini and Qwen2.5-7B, and $\approx$2--6 for the weaker models; the ranking is unchanged.
Prompt sensitivity (T.A3): The same fine-tuned model scores Ans 8.44 with a minimal tool schema and 45.26 with the full one.
Disaster tasks make up $\approx$17\% of the test set, but 2.7\% of the training set (Fig.~4a).

\begin{table}[h]
\centering\scriptsize\setlength{\tabcolsep}{3pt}
\setlength{\aboverulesep}{0pt}\setlength{\belowrulesep}{0pt}
\rowcolors{2}{gray!20}{gray!2}
\begin{tabular}{lcccc}
\toprule
\rowcolor{gray!50} \multicolumn{1}{c}{\textbf{Model}} & \textbf{Tool} & \textbf{ArgV} & \textbf{GIS F1} & \textbf{Ans} \\
\midrule
GPT-5 & 82.69 & 37.28 & 91.50 & 43.88 \\
GPT-4o & 93.88 & 45.80 & 95.80 & 39.22 \\
o4-mini & 68.33 & 37.95 & 86.51 & 35.18 \\
Qwen2.5-7B & 85.51 & 38.00 & 75.46 & 15.85 \\
Llama-3.1-8B & 39.30 & 17.49 & 64.64 & 12.70 \\
InternLM3-8B & 38.46 & 13.25 & 31.47 & 0.24 \\
Mistral-7B & 35.12 & 12.45 & 17.12 & 0.39 \\
Qwen2.5-3B & 72.13 & 24.12 & 35.64 & 1.83 \\
Qwen3-4B & 86.94 & 33.55 & 71.82 & 13.72 \\
OEA-4B (fine-tuned) & 97.18 & 62.10 & 98.52 & 45.26 \\
\bottomrule
\end{tabular}
\caption{OpenEarthAgent (T.3 and T.4 of~\cite{shabbir2026openearthagent}).}
\label{tab:a-oea}
\end{table}

\para{Earth-Bench~\cite{feng2026earthagent}} (248 questions posed in auto-planning (AP) and instruction-following (IF) regimes; 104 tools; 13 models; deterministic exact match on multiple-choice answers).
Best accuracy GPT-5 65.99 (AP) / 62.35 (IF).
Parameter accuracy $\leq$ 30.81 (AP) and $\leq$ 31.36 (IF) for all models, while tool-any-order recall reaches 78.31 (AP) and 78.86 (IF).
Spearman $\rho$ with accuracy ($n=13$, AP / IF): Tool-any-order 0.68 / 0.78, tool-in-order 0.70 / 0.78, exact match 0.63 / 0.70, parameters 0.64 / 0.67 (all $p<0.05$).
The model with the best trajectories (DeepSeek-V3.1) ranks fourth in accuracy (51.42).
LLaMA-4 reaches 44.94 (AP) with tool-any-order 16.51 and parameter accuracy 1.30, above GPT-4o (43.72, parameters 26.55); InternVL-3.5 scores 26.72 with tool-any-order 8.83, close to the 25\% chance level of the four-option examples; tool-free LLMs reach about 37\% (Fig.~6 text).
Step guidance (IF) leaves accuracy equal or lower for 6 of 13 models, although tool-any-order rises for all 13.
Accuracy by number of tool calls (T.12): GPT-5 81--90\% at 3--4 calls, 0/3 at 19 calls.
Error types (Fig.~7): Invalid parameters are 81.8\% of GPT-5's errors (AP); file and tool hallucinations plus unawareness of termination are 57--78\% of errors for Kimi-K2, Qwen3-Max and DeepSeek-V3.1.

\para{DORA~\cite{wang2026dora}} (515 tasks from 45 disaster events of 10 hazard types; 108 tools; 8 modalities; 13 agents; deterministic typed tolerances).
\cref{tab:a-dora} lists final accuracy and trajectory metrics (T.4).
The top six agents span 51.17--53.74 (2.57\pp).
Gold trajectory: 80.48 with all trajectory metrics at 100.
Gold tool-order hints (T.5, IF \vs AP): Exact match $+26.98$ / $+28.94$ / $+25.63$\pp{} and final accuracy $+1.97$ / $+1.08$ / $+4.40$\pp{} for Gemini-3.0-Flash / Gemma-4-31B / MiniMax-M2.7.
Gold minus best agent by gold length (Fig.~9): 2--3 calls 7\% ($n=109$), 4--5 22\% (150), 6--7 19\% (92), 8--10 34\% (94), $\geq$11 56\% (70); growth is not monotonic.
Failure taxonomy (Fig.~7; 91.4\% classified): Composition 56.3 (wrong upstream reference 38.5, missing spatial tools 12.8, wrong phenomenon tool 3.5, length/area confusion 1.5), damage-semantic grounding 20.3 (encoding evolution 14.3, term mismatch 6.0), sensor--modality mismatch 14.8 (raster for vector 9.1, other 5.7).
Modality heatmap (Fig.~8), Gemini-3.0-Flash: Optical 59.22, +vector 27.91, +SAR 34.09, MS+DEM+slope 17.27, +DSM+thermal 53.44; 7 of 13 agents score exactly 17.27 on MS+DEM+slope (maximum 21.82).
Scaffolds (T.6, Gemma-4-31B): Plan-then-execute 54.41 at 2.2$\times$ ReAct latency; ReWOO 44.26.

\begin{table}[h]
\centering\scriptsize\setlength{\tabcolsep}{3pt}
\setlength{\aboverulesep}{0pt}\setlength{\belowrulesep}{0pt}
\rowcolors{2}{gray!20}{gray!2}
\begin{tabular}{lcccc}
\toprule
\rowcolor{gray!50} \multicolumn{1}{c}{\textbf{Agent}} & \textbf{Final} & \textbf{T-Any-O} & \textbf{T-Exact} & \textbf{ParAcc} \\
\midrule
Gold trajectory & 80.48 & 100 & 100 & 100 \\
Gemini-3.0-Flash & 53.74 & 75.17 & 35.55 & 42.63 \\
Qwen3.5-397B-A17B & 53.45 & 75.44 & 32.13 & 36.50 \\
MiMo-V2-Pro & 52.89 & 74.05 & 34.26 & 38.45 \\
Grok-4.1-Fast & 52.10 & 59.15 & 28.09 & 31.96 \\
Claude-Sonnet-4.6 & 52.01 & 76.29 & 31.43 & 39.48 \\
Gemma-4-31B & 51.17 & 69.89 & 32.82 & 38.23 \\
MiniMax-M2.7 & 48.35 & 62.89 & 25.32 & 29.68 \\
DeepSeek-V3.2 & 48.23 & 74.26 & 30.39 & 34.69 \\
GPT-5.4 & 47.63 & 66.84 & 31.82 & 37.71 \\
Step-3.5-Flash & 46.68 & 62.59 & 25.33 & 29.43 \\
GPT-5.4-Nano & 38.14 & 54.96 & 21.81 & 27.02 \\
GPT-OSS-120B & 35.11 & 48.69 & 22.79 & 26.24 \\
Qwen3.5-35B-A3B & 24.01 & 55.20 & 23.10 & 27.08 \\
Gemini-3.0-Flash, no tools & 18.55 & -- & -- & -- \\
\bottomrule
\end{tabular}
\caption{DORA (T.4 of~\cite{wang2026dora}), sorted by final accuracy.}
\label{tab:a-dora}
\end{table}

\para{UnivEARTH~\cite{kao2026univearth}} (408 yes/no questions over Google Earth Engine data; 8 models; a Gemini-2.5-Flash call extracts the answer from the script output; no unanswerable questions, per the Limitations section).
\cref{tab:a-univearth} reproduces T.3.
``Code fails over 44\% of the time'' refers to all questions without a valid answer (Claude: C1+C2+C3+D $=$ 44.2); only category D (21.4) are crashes.
Reflexion (Fig.~5; correct / wrong at rounds 0, 1, 2): Claude-4.5-Sonnet 40.0/15.7, 58.1/23.8, 64.5/26.0; Gemini-2.5-Pro 36.8/9.6 (36.9 in T.3), 52.7/12.7, 59.8/15.7; GPT-5 33.1/12.0, 53.7/18.9, 59.3/21.8.
Precision: 0.72$\to$0.71$\to$0.71, 0.79$\to$0.81$\to$0.79, 0.73$\to$0.74$\to$0.73.
Share of newly produced answers that are wrong: 29.6\%, 21.0\%, 27.2\%.
Providing the correct collection IDs (Fig.~4 and Sec.~4.3, VIIRS) raises Claude's accuracy 38.2$\to$61.5, but its wrong answers 8$\to$23.1\%; OSCAR re-planning (T.6) lowers Gemini-2.5-Pro's precision 0.79$\to$0.72.

\begin{table}[h]
\centering\scriptsize\setlength{\tabcolsep}{3pt}
\setlength{\aboverulesep}{0pt}\setlength{\belowrulesep}{0pt}
\rowcolors{2}{gray!20}{gray!2}
\begin{tabular}{lcccc}
\toprule
\rowcolor{gray!50} \multicolumn{1}{c}{\textbf{Model}} & \textbf{Correct} & \textbf{Wrong} & \textbf{No valid answer} & \textbf{Precision} \\
\midrule
Claude-4.5-Sonnet & 40.0 & 15.7 & 44.2 & 0.72 \\
Gemini-2.5-Pro & 36.9 & 9.6 & 53.5 & 0.79 \\
GPT-5 & 33.1 & 12.0 & 54.9 & 0.73 \\
DeepSeek-Reasoner & 28.5 & 13.0 & 58.6 & 0.69 \\
DeepSeek-Chat & 27.0 & 13.0 & 60.0 & 0.68 \\
Kimi-k2 & 17.0 & 8.4 & 74.7 & 0.67 \\
Claude-4.5-Haiku & 16.7 & 12.8 & 70.5 & 0.57 \\
Gemini-2.5-Flash & 10.6 & 5.9 & 83.4 & 0.64 \\
\bottomrule
\end{tabular}
\caption{UnivEARTH zero-shot (T.3 of~\cite{kao2026univearth}); precision $=$ correct/(correct$+$wrong) is derived.}
\label{tab:a-univearth}
\end{table}

\para{GeoBenchX~\cite{krechetova2025geobenchx}} (202 tasks: 123 solvable, 79 unsolvable, with a \texttt{reject\_task} tool; 23 tools; 8 analyzed models, single run; a panel of three LLM judges compares solutions with references; judges match human scores on 96/94/88\% of 50 tasks; final numeric answers are not evaluated).
\cref{tab:a-geobenchx} reproduces T.1--T.2 and T.8.
$\rho$(solvable, rejection) $=-0.25$ ($p=0.55$, $n=8$); $-0.63$ ($p=0.13$) without o4-mini.
o4-mini's rejection score does not penalize false rejections: It rejects 49\% of all tasks (Sec.~6.3), \ie $\approx$99 rejections for $\approx$71 correct ones, so $\approx$28 solvable tasks ($\approx$23\%) are rejected.

\begin{table}[h]
\centering\scriptsize\setlength{\tabcolsep}{3pt}
\setlength{\aboverulesep}{0pt}\setlength{\belowrulesep}{0pt}
\rowcolors{2}{gray!20}{gray!2}
\begin{tabular}{lcccc}
\toprule
\rowcolor{gray!50} \multicolumn{1}{c}{\textbf{Model}} & \textbf{Solvable} & \textbf{Rejection} & \textbf{In (M)} & \textbf{Out (k)} \\
\midrule
Claude Sonnet 4 & 0.55 & 0.17 & 26.9 & 291.6 \\
o4-mini & 0.51 & 0.90 & 7.9 & 956.6 \\
Claude Sonnet 3.5 & 0.47 & 0.47 & 12.2 & 151.2 \\
GPT-4.1 & 0.42 & 0.43 & 12.7 & 121.3 \\
Claude Haiku 3.5 & 0.40 & 0.22 & 14.9 & 196.0 \\
Gemini 2.0 Flash & 0.39 & 0.40 & 7.5 & 40.7 \\
Gemini 2.5 Pro Prev. & 0.39 & 0.55 & 10.4 & 85.4 \\
GPT-4o & 0.38 & 0.62 & 8.0 & 61.2 \\
\bottomrule
\end{tabular}
\caption{GeoBenchX match rates and token use (T.1, T.2, T.8 of~\cite{krechetova2025geobenchx}).}
\label{tab:a-geobenchx}
\end{table}

\para{GeoNatureAgent~\cite{diazireland2026geonatureagent}} (93 tasks in 18 categories; 16 tools; 9 models $\times$ 3 seeds; eight deterministic checks per case, no LLM judge; deliberately unsolvable error-handling tasks, count not reported).
\cref{tab:a-gna} reproduces T.1 and the error-handling column of Fig.~4.
The $\pm$ statistic is not defined in the paper; its mean is 2.37\pp{} (max 4.3).
Adjacent rank pairs closer than the larger of their two spreads: GLM-5/Gemini 2.5 Pro (2.2 \vs 3.3) and GPT-4o/Qwen3-235B (0.4 \vs 4.3), \ie 2 of 8.
$\rho$(accuracy, error handling) $=0.33$ ($p=0.38$, $n=9$).
DeepSeek V3.2 reaches 56.3/60.8 $=$ 93\% of Claude Sonnet 4's accuracy at \$1.05 \vs \$11.84 per 93-case run (11.3$\times$; 11.6$\times$ from the rounded per-case costs).
Partial-credit scores (66--90) far exceed binary accuracy (16--61): Many failures are near-misses that a strict check rejects.

\begin{table}[h]
\centering\scriptsize\setlength{\tabcolsep}{3pt}
\setlength{\aboverulesep}{0pt}\setlength{\belowrulesep}{0pt}
\rowcolors{2}{gray!20}{gray!2}
\begin{tabular}{lccc}
\toprule
\rowcolor{gray!50} \multicolumn{1}{c}{\textbf{Model}} & \textbf{Accuracy} & \textbf{Error handling} & \textbf{\$ per run} \\
\midrule
Claude Sonnet 4 & 60.8 $\pm$ 0.8 & 50 & 11.84 \\
DeepSeek V3.2 & 56.3 $\pm$ 3.1 & 44 & 1.05 \\
GLM-5 & 50.2 $\pm$ 2.2 & 44 & 3.58 \\
Gemini 2.5 Pro & 48.0 $\pm$ 3.3 & 72 & 4.87 \\
GPT-4o & 41.6 $\pm$ 2.7 & 44 & 6.49 \\
Qwen3-235B & 41.2 $\pm$ 4.3 & 44 & 0.89 \\
GPT-OSS-120B & 34.1 $\pm$ 1.2 & 33 & 8.27 \\
Llama 4 Scout & 26.9 $\pm$ 2.1 & 39 & 0.32 \\
Gemma-3-27B & 15.8 $\pm$ 1.6 & 50 & 5.81 \\
\bottomrule
\end{tabular}
\caption{GeoNatureAgent (T.1 and Fig.~4 of~\cite{diazireland2026geonatureagent}).}
\label{tab:a-gna}
\end{table}

\para{Geo-OLM~\cite{stamoulis2025geoolm}} (2{,}000 queries; success $=$ fully completed tasks, correctness $=$ correct calls with correct parameters in the expected order; total \$ per 2k queries).
\cref{tab:a-geoolm} reproduces the GPT-4o rows of T.1.
Success--correctness gaps: 8.49 to 54.30\pp{} for GPT-4o and 15.69 to 54.25\pp{} for GPT-4o-mini across frameworks.
Cost per unit of correctness $=$ total\,\$ $/$ (2000 $\times$ correctness): A derived proxy, because correctness is a per-call fraction.
Self-reflection can harm: Llama-3.3-70B with GeoLLM-Engine drops from 41.33 to 4.37 correctness when error/termination reflection is added (T.3).
In the Turkey-earthquake case study (T.5, 120 questions), success is 100\% for every agent with GPT-4o and GPT-4o-mini, while correctness ranges 58.4--95.9.
Qwen-2.5-7B with Geo-OLM reaches 63.3 success and 40.9 correctness for \$0.15, an estimated 513$\times$ cheaper than GPT-4o with Geo-OLM (open-model costs are estimated from GPU rates).

\begin{table}[h]
\centering\scriptsize\setlength{\tabcolsep}{3pt}
\setlength{\aboverulesep}{0pt}\setlength{\belowrulesep}{0pt}
\rowcolors{2}{gray!20}{gray!2}
\begin{tabular}{lcccc}
\toprule
\rowcolor{gray!50} \multicolumn{1}{c}{\textbf{Framework (GPT-4o)}} & \textbf{Success} & \textbf{Correct.} & \textbf{Total \$} & \textbf{\$/corr.\ unit} \\
\midrule
GeoLLM-QA & 62.3 & 52.49 & 68.23 & 0.065 \\
GeoLLM-Engine & 82.6 & 74.11 & 64.72 & 0.044 \\
Chameleon & 55.1 & 21.87 & 115.08 & 0.263 \\
Magentic & 79.7 & 25.40 & 447.55 & 0.881 \\
GeoLLM-Squad & 96.2 & 85.32 & 219.67 & 0.129 \\
Geo-OLM & 95.1 & 86.03 & 77.02 & 0.045 \\
GeoLLM-Squad (o1) & 97.2 & 90.97 & 833.91 & 0.458 \\
\bottomrule
\end{tabular}
\caption{Geo-OLM benchmark, GPT-4o and o1 rows (T.1 of~\cite{stamoulis2025geoolm}); last column derived.}
\label{tab:a-geoolm}
\end{table}

\para{GeoLLM-Engine~\cite{singh2024geollmengine}} (main evaluation on 10{,}000 queries / 50{,}830 tasks; 2 LLMs $\times$ 3 agent methods $\times$ zero/few-shot; reference solutions accepted when 9 of 10 GPT-4 runs agree).
Correctness 38.1--84.3 and success 40.9--81.1 across the 12 configurations (T.4); success exceeds correctness by up to 24.5\pp{} (GPT-3.5 CoT zero-shot), so success does not certify a valid process.
GPT-4 Turbo error composition (Fig.~5): Missed function 56.6--63.2\%, argument error 21.5--24.5\%, function error 14.3--20.9\%.
Success with one tool is 100\%, falling to $\approx$0.32 (CoT zero-shot) to $\approx$0.70 (ReAct zero-shot) with ten tools (Fig.~4).
Tokens per task: 17.3k--40.7k; token use does not correlate with success ($\rho=-0.31$, $p=0.33$, $n=12$).

\para{GeoLLM-Squad~\cite{lee2025geollmsquad}} (2{,}000 generated tasks from 56 human-vetted seeds; GPT-4o-mini and two Qwen-2.5 models; correctness $=$ share of correct tool steps in order).
Multi-agent Squad reaches 60.29 correctness \vs 43.32 for the single-agent baseline at 3.5$\times$ the tokens (78.5k \vs 22.2k); Magentic uses 141.7k--210.1k tokens for 30.1--34.0 correctness (T.3).
With Qwen-2.5-3B, correctness drops to 36.95 (Squad), 21.11 (Chameleon) and 7.81 (Magentic) (T.4).

\para{Spatial-Agent~\cite{bao2026spatialagent}} (MapEval-API and a MapQA subset; map and POI APIs, OpenStreetMap data, no imagery; accuracy on multiple-choice questions, protocol not detailed; single run).
Question counts are not stated; the category percentages imply 280 answerable MapEval-API questions and at least 267 MapQA questions. Its reported overall scores are unweighted means of the category scores, except for GPT-5 (71.88 \vs 73.08) and GPT-3.5-Turbo (34.61 \vs 36.51).
Direct, ReAct and Reflexion baselines are reported only for GPT-4o-mini (\cref{tab:a-spatial}).
On MapEval-API, scaffold and tool set move GPT-4o-mini from 23.00 (MapEval's ReAct agent, per question over all 300 questions; 24.88 as the mean of its answerable-category scores) to 32.98 (ReAct), 38.29 (Reflexion), 41.17 (Plan-and-Solve, T.3) and 45.15 (Spatial-Agent), while tool-free prompting (Direct) reaches 32.23; ReAct falls below Direct on the Nearby ($-9.6$\pp) and Trip ($-3.0$\pp) categories.
With Spatial-Agent fixed, backbones span 22.70 (Gemma-2-9B) to 71.88 (GPT-5).
Error analysis of 68 failures (Fig.~3; backbone not stated): Data-quality issues of the external APIs 45.6\%, search-result mismatch 33.8\%, concept and role assignment 10.3\%, wrong answer after correct execution 10.3\%, and none from plan (graph) construction.
Cost (GPT-4o-mini, \S4.3): Tokens per query 1{,}500/50 (Direct), 7{,}354/735 (ReAct), 10{,}964/1{,}271 (Reflexion) and 9{,}185/1{,}451 (Spatial-Agent), all below \$0.003 per query; latency 0.6\,s (Direct) \vs 7.5--15.4\,s for the agents (Fig.~4).
Derived \$ per correct MapEval-API answer: Direct \$0.0008, ReAct \$0.0047, Spatial-Agent \$0.0050, Reflexion \$0.0063 (cost per query divided by accuracy; token counts assumed identical across datasets).
No repeated runs, significance tests, unanswerable questions or confidence estimates are reported.

\begin{table}[h]
\centering\scriptsize\setlength{\tabcolsep}{3pt}
\setlength{\aboverulesep}{0pt}\setlength{\belowrulesep}{0pt}
\rowcolors{2}{gray!20}{gray!2}
\begin{tabular}{lccc}
\toprule
\rowcolor{gray!50} \multicolumn{1}{c}{\textbf{Method}} & \textbf{Backbone} & \textbf{MapEval-API} & \textbf{MapQA} \\
\midrule
Direct (no tools) & GPT-4o-mini & 32.23 & 13.55 \\
MapEval ReAct agent & GPT-4o-mini & 23.00 & -- \\
ReAct & GPT-4o-mini & 32.98 & 43.79 \\
Reflexion & GPT-4o-mini & 38.29 & 53.79 \\
Spatial-Agent & GPT-4o-mini & 45.15 & 61.45 \\
Spatial-Agent & GPT-3.5-Turbo & 34.61 & -- \\
Spatial-Agent & GPT-5 & 71.88 & -- \\
Spatial-Agent & LLaMA-70B & 47.77 & 62.45 \\
Spatial-Agent & Qwen2.5-72B & 53.41 & 61.45 \\
Spatial-Agent & Qwen2.5-32B & 52.35 & -- \\
Spatial-Agent & Gemma-2-9B & 22.70 & -- \\
\bottomrule
\end{tabular}
\caption{Spatial-Agent (T.1 and T.2 of~\cite{bao2026spatialagent}); reported overall accuracy, except the MapEval row (per question over 300 questions).}
\label{tab:a-spatial}
\end{table}

\para{MapEval~\cite{dihan2025mapeval}} (700 multiple-choice questions over 180 cities; the agentic MapEval-API track poses the 300 textual questions without context to zero-shot ReAct agents with five Google-Maps-based tools over a cached environment; 11 models; deterministic multiple-choice scoring; single run).
Twenty questions per track (6.7\% of MapEval-API) are unanswerable, so each counts 5\pp{} in that column.
\cref{tab:a-mapeval} reproduces T.4 with the textual scores of T.3 for the same models.
All 11 models score lower on MapEval-API than on MapEval-Textual (same questions, context given instead of retrieved): Mean $-16.5$\pp, range $-2.3$ to $-28.0$\pp; the rank order is largely preserved ($\rho=0.84$, $p=0.001$).
Unanswerable accuracy (API) ranges 5--65\%; the most accurate agent (Claude-3.5-Sonnet, 64.0) abstains correctly on 55\%, Gemini-1.5-Pro (43.3, fourth) on 65\%, GPT-4o on 15\%.
With answerable accuracy $=(300\cdot\text{overall}-20\cdot\text{unans.})/280$, $\rho$(answerable, unanswerable) $=0.50$ ($p=0.12$, $n=11$).
Moving from textual to API access lowers unanswerable accuracy for 7 of 11 models.
Agents hit the iteration limit 90 times in 3{,}300 runs (GPT-3.5-Turbo 16; Fig.~19, read on grid lines).
Changing the agent from ReAct to Chameleon raises GPT-3.5-Turbo from 27.33 to 49.33 (T.7).
In the open-ended ablation (App.~G.2), an o3-mini judge credits Claude with 55\% on unanswerable textual questions, where manual inspection finds $\approx$80\% correct.
The human baseline (86.67 overall, 65 unanswerable) is measured on the textual track, reported as the better of two annotators, and reused for the API track.

\begin{table}[h]
\centering\scriptsize\setlength{\tabcolsep}{3pt}
\setlength{\aboverulesep}{0pt}\setlength{\belowrulesep}{0pt}
\rowcolors{2}{gray!20}{gray!2}
\begin{tabular}{lccc}
\toprule
\rowcolor{gray!50} \multicolumn{1}{c}{\textbf{Model}} & \textbf{API overall} & \textbf{API unans.} & \textbf{Textual overall} \\
\midrule
Claude-3.5-Sonnet & 64.00 & 55 & 66.33 \\
GPT-4-Turbo & 53.67 & 25 & 62.33 \\
GPT-4o & 48.67 & 15 & 63.33 \\
Gemini-1.5-Pro & 43.33 & 65 & 66.33 \\
Gemini-1.5-Flash & 41.67 & 30 & 58.67 \\
Llama-3.2-90B & 39.67 & 15 & 58.33 \\
Llama-3.1-70B & 37.67 & 15 & 61.00 \\
Mixtral-8x7B & 27.67 & 15 & 43.00 \\
GPT-3.5-Turbo & 27.33 & 15 & 37.67 \\
Gemma-2-9B & 27.00 & 45 & 47.33 \\
GPT-4o-mini & 23.00 & 5 & 51.00 \\
Human (textual) & 86.67 & 65 & 86.67 \\
\bottomrule
\end{tabular}
\caption{MapEval-API (T.4) and MapEval-Textual (T.3) of~\cite{dihan2025mapeval}.}
\label{tab:a-mapeval}
\end{table}

\subsection{Insights behind findings F1--F6}
\label{app:insights}

\begin{enumerate}[label=I\arabic*.,leftmargin=2.2em]
\item \emph{The ecosystem does not measure trust} (F1): 1/12 repeated runs, 3/12 unanswerable tasks, 3/12 dollar cost, 0/12 calibration, 4/12 LLM judges, 3/12 LLM-generated references.
\item \emph{Right tools, wrong answers} (F2): ThinkGeo GPT-4o 67.7 \vs 9.8; OpenEarthAgent GPT-4o 93.9 \vs 39.2; tool--answer $\rho=0.40$ (n.s.) and 0.70.
\item \emph{Completed is not correct} (F2): Geo-OLM gaps 8.5--54.3\pp; GeoLLM-Engine up to 24.5\pp; DORA gold trajectory 80.5; Spatial-Agent 10.3\% of failures after correct execution; Earth-Bench answers without grounded parameters.
\item \emph{Scoring changes gap sizes} (F6): Lenient judge $+\approx$15 \vs $+\approx$7--9.5 points (OpenEarthAgent); judge--human agreement 88--96\% (GeoBenchX); LLM judge 55\% \vs $\approx$80\% by manual inspection (MapEval); same backbone $\approx$24.9--45.2\% across five scaffolds and tool sets (Spatial-Agent, MapEval).
\item \emph{Model gaps within noise} (F4): DORA top six within 2.6\pp; GeoNatureAgent spread $\pm$2.4\pp{} and 2/8 adjacent pairs unresolved.
\item \emph{Retries add coverage, not reliability} (F5): UnivEARTH precision flat, 21--30\% of new answers wrong; Geo-OLM self-reflection can collapse correctness.
\item \emph{Capability and honesty are different axes} (F5): GeoBenchX $\rho=-0.25$; GeoNatureAgent $\rho=0.33$; MapEval-API $\rho=0.50$ (all n.s.); best-accuracy models are not best at rejection.
\item \emph{Grounding, not planning} (F3): DORA gold-order hints $+25.6$--$28.9$\pp{} exact match but $+1.1$--$4.4$\pp{} answers; Earth-Bench parameters $\leq$31.4; failure taxonomies dominated by wiring, semantics, sensor mismatch and faulty retrieved data (Spatial-Agent 79.4\%); retrieving context through map APIs costs all 11 MapEval agents accuracy (mean $-16.5$\pp).
\item \emph{Reliability decays with horizon and crisis complexity} (F4): DORA gap 7$\to$56\pp; modality drop to 17.3; OpenEarthAgent disaster domain 21.8--24.9; ThinkGeo SAR $\leq$5.6.
\item \emph{Cost per correct answer is not reported} (F6): Derived \$0.044--\$0.881 per unit of correctness (Geo-OLM); 93\% of accuracy at $\approx$1/11 of the cost (GeoNatureAgent); tool use multiplies the cost per correct answer by $\approx$6--8 (Spatial-Agent).
\end{enumerate}

\subsection{Limitations of the current analysis}
\label{app:limits}

(i)~Tasks, data and scoring differ across benchmarks, so only within-paper contrasts are interpreted, and model generations range from GPT-3.5 (2024) to GPT-5.4 and Gemini-3 (2026).
(ii)~All numbers originate from the benchmark papers themselves, not from independent re-runs.
(iii)~Correlations use 8--13 models and are descriptive.
(iv)~A few values are taken from figures: Printed labels (DORA Fig.~7--9, UnivEARTH Fig.~5, OpenEarthAgent Fig.~A3, GeoNatureAgent Fig.~4, GeoLLM-Engine Fig.~5, Spatial-Agent Fig.~3--4), bars read on grid lines (MapEval Fig.~19) and plotted positions marked $\approx$ (OpenEarthAgent Fig.~A7, GeoLLM-Engine Fig.~4).
(v)~Crisis-specific evidence is concentrated in DORA; the other benchmarks contain disaster tasks as a minority or not at all, which is itself a gap for the community to close.
(vi)~MapEval and Spatial-Agent test map and POI tools without imagery; their evidence is used for tool use, abstention, scaffolds and cost, not for EO perception.
(vii)~Three studies are agent-method papers on shared task sets (Geo-OLM and GeoLLM-Squad within the GeoLLM-Engine environment, Spatial-Agent on MapEval-API), so their results are not independent of the benchmarks they reuse.

\end{document}